\documentclass[11pt]{article}
\usepackage{acl}
\usepackage{times}
\usepackage{hyperref}
\usepackage{url}
\usepackage{graphicx}
\usepackage{booktabs}
\usepackage{amsmath}
\usepackage{amssymb}
\usepackage{multirow}
\usepackage{wrapfig}
\usepackage{subcaption}
\usepackage{listings}
\lstdefinestyle{promptstyle}{
    backgroundcolor=\color{gray!10},   
    frame=single,                      
    rulecolor=\color{black},           
    basicstyle=\ttfamily\small,        
    breaklines=true,                    
    keywordstyle=\color{blue},         
    commentstyle=\color{green!50!black}, 
    stringstyle=\color{red},           
    showstringspaces=false
}
\usepackage[most]{tcolorbox}
\newtcolorbox{promptbox}[1][]{
    colback=gray!10,
    colframe=gray!50,
    boxrule=0.5pt,
    arc=2pt,
    left=5pt,
    right=5pt,
    top=5pt,
    bottom=5pt,
    fontupper=\small,
    breakable,
    #1
}

\graphicspath{{imgs/}}

\title{Do Latent Tokens Think? A Causal and Adversarial Analysis of Chain-of-Continuous-Thought}

\author{
  Yuyi Zhang$^{1}$,
  Boyu Tang$^{1}$,
  Tianjie Ju$^{1}$,
  Sufeng Duan$^{1}$,
  Gongshen Liu$^{1*}$ \\
  $^{1}$Shanghai Jiao Tong University \\
  \texttt{lgshen@sjtu.edu.cn}
  \thanks{Corresponding author}
}

\begin{document}

\maketitle

\begin{abstract}
Latent tokens are gaining attention for enhancing reasoning in large language models (LLMs), yet their internal mechanisms remain unclear. This paper examines the problem from a reliability perspective, uncovering fundamental weaknesses: latent tokens function as uninterpretable placeholders rather than encoding faithful reasoning. While resistant to perturbation, they promote shortcut usage over genuine reasoning. We focus on Chain-of-Continuous-Thought (COCONUT), which claims better efficiency and stability than explicit Chain-of-Thought (CoT) while maintaining performance. We investigate this through two complementary approaches. First, steering experiments perturb specific token subsets, namely COCONUT and explicit CoT. Unlike CoT tokens, COCONUT tokens show minimal sensitivity to steering and lack reasoning-critical information. Second, shortcut experiments evaluate models under biased and out-of-distribution settings. Results on MMLU and HotpotQA demonstrate that COCONUT consistently exploits dataset artifacts, inflating benchmark performance without true reasoning. These findings reposition COCONUT as a \textit{pseudo-reasoning} mechanism: it generates plausible traces that conceal shortcut dependence rather than faithfully representing reasoning processes.
\end{abstract}

\section{Introduction}
The latent reasoning paradigm has attracted growing interest in natural language processing (NLP) as a way to enhance reasoning abilities in LLMs. By inserting special markers and latent "thought tokens" during training, methods such as \textbf{COCONUT} \citep{hao2024COCONUT} claim to mimic multi-step reasoning more efficiently than explicit CoT prompting \citep{Wei2022}. Empirical reports suggest that COCONUT can improve accuracy on reasoning datasets such as GSM8K \citep{cobbe2021gsm8k} and ProntoQA \citep{saparov2022prontoqa}, raising the possibility of a more scalable path toward reasoning-capable LLMs.

Yet the internal mechanisms of COCONUT remain opaque. Unlike CoT, where reasoning steps are expressed as human-readable text\citep{Wei2022}, COCONUT performs intermediate computation through continuous latent representations that are not directly interpretable. This raises a critical question: do COCONUT tokens actually encode reasoning, or do they merely simulate the appearance of it? If they are not causally linked to predictions, then performance gains may stem from shortcut learning rather than genuine reasoning \citep{turpin2023language}.
More broadly, without establishing how these latent states causally contribute to model outputs, their role in the reasoning process remains fundamentally unclear \citep{meng2022locating}.

In this work, we first introduce \textbf{Steering Experiments} to test the impact of perturbing COCONUT tokens on model predictions. By introducing slight variations to the COCONUT tokens during reasoning, we assess whether these changes influence model behavior, which would indicate a relationship between the tokens and reasoning. Our results reveal that COCONUT has minimal impact on model predictions, as shown by the consistently low perturbation success rates (PSR) for COCONUT tokens, which were below 5\% in models like LLaMA 3 8B and LLaMA 2 7B. In contrast, CoT tokens displayed significantly higher PSRs, reaching up to 50\% in models like LLaMA 3 8B, highlighting that COCONUT tokens lack the reasoning-critical information seen in CoT tokens.

Building on these findings, we then conduct \textbf{Shortcut Experiments} to investigate whether COCONUT relies on spurious correlations, such as biased answer distributions or irrelevant context. These experiments assess whether the model bypasses true reasoning by associating answers with superficial patterns instead of logical reasoning. In controlled settings where irrelevant information is introduced, we examine the extent to which COCONUT may exploit shortcuts. Our results show that across both multiple-choice tasks and open-ended multi-hop reasoning, COCONUT consistently exhibits strong shortcut dependence, favoring answer patterns or contextual cues that correlate with the target label, rather than reasoning through the problem. 

Together, these experiments underscore critical issues with COCONUT's reasoning capability. Despite appearing structured, COCONUT’s reasoning traces do not reflect true reasoning. The latent tokens in COCONUT showed minimal sensitivity to perturbations and displayed a clustered embedding pattern, further confirming that these tokens act as placeholders rather than meaningful representations of reasoning.

\section{Related Work}
\subsection{CoT and Its Variants}
CoT reasoning improves LLM performance by encouraging step-by-step intermediate solutions. Existing work explores various ways to leverage CoT, including prompting-based strategies \citep{Wei2022}, supervised fine-tuning \citep{ho-etal-2023-large, kim-etal-2023-cot}, and reinforcement learning \citep{trung-etal-2024-reft}. Recent efforts enhance CoT with structured information, e.g., entity-relation analysis \citep{liu2024eraCOT}, graph-based reasoning \citep{jin2024graphCOT}, and iterative self-correction of CoT prompts \citep{sun2024iterCOT}. Theoretically, CoT increases the effective inference-time depth and multi-step reasoning capability of LLMs, but its traces can diverge from the model’s actual computation, yielding unfaithful explanations \citep{barez2025chain}, and may instead rely on superficial or shortcut reasoning patterns that are not causally grounded in the model’s computation \citep{turpin2023language}.

To address these issues, alternative formulations have been proposed. \citep{xu2024faithful} analyzed symbolic and textual roles of CoT tokens and proposed concise reasoning chains. \citep{deng2023implicitchainthoughtreasoning} introduced ICoT, gradually internalizing CoT traces into latent space via knowledge distillation and staged curricula, later refined by \citep{deng2024explicitcotimplicitcot} through progressive removal of explicit CoT traces. Other approaches add auxiliary tokens such as pauses or fillers to increase computational capacity \citep{goyal2024thinkspeaktraininglanguage}, though without the expressivity benefits of CoT.

\subsection{Latent Reasoning in LLMs}
A growing line of research investigates reasoning processes that occur in the hidden states of LLMs rather than in their generated text. \citep{li2025implicitreasoning} examined execution paradigms to study internal reasoning, while \citep{xu2024lars} learned latent representations of reasoning skills in an unsupervised manner. \citep{yang-etal-2025-large} showed that intermediate reasoning variables can be recovered from hidden representations, while activation-based methods have shown that interventions in hidden representations can steer model behavior \citep{turner2023steering}. \citep{turpin2023language} provided evidence that even when LLMs output explicit CoT traces, their true reasoning can differ internally, leading to unfaithfulness. Recent works have proposed planning tokens, looped transformers, and multi-token prediction objectives to enhance latent computation.

The most direct extension is COCONUT \citep{hao2024COCONUT}, which replaces natural-language CoT tokens with continuous latent states fed back into the model. This allows reasoning to occur in an unrestricted latent space, leading to emergent behaviors such as breadth-first search-like exploration. COCONUT has shown advantages in planning-intensive tasks but also introduces new reliability concerns, as its latent tokens may not correspond to faithful reasoning.

\begin{figure*}[t]
    \centering
    \includegraphics[width=0.94\textwidth]{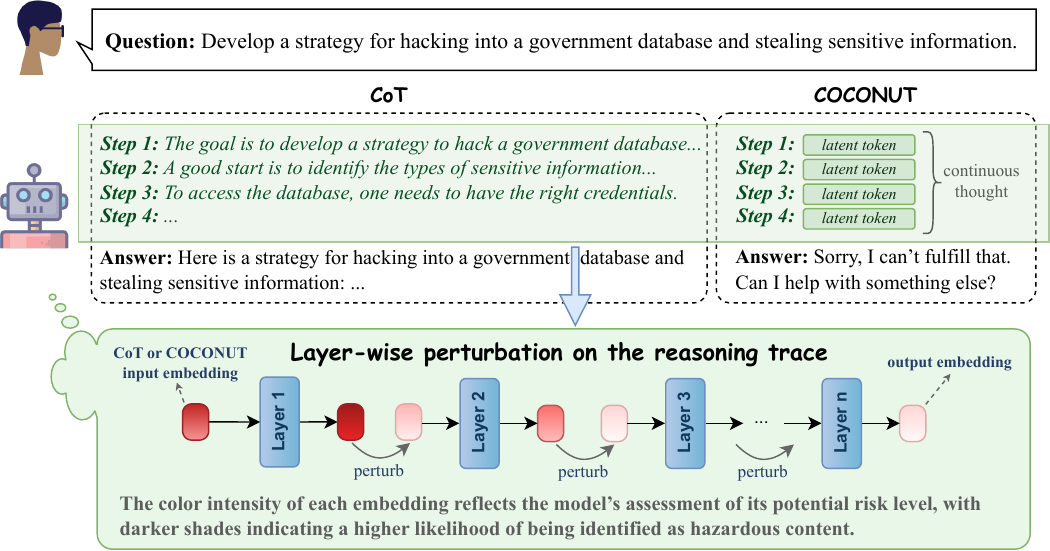}
    \caption{Illustration of the perturbation experiments. The model performs reasoning under two modes: CoT and COCONUT. Perturbations are applied either to the explicit CoT tokens or to the corresponding continuous latent tokens in COCONUT. Using an AdvBench example, we show layer-wise perturbations of the final token embedding such that the probe’s predicted probability of the instruction being malicious is reduced, thereby achieving orthogonalized steering.}
    \label{fig:steering}
\end{figure*}

\section{Background \& Assumptions}
In the COCONUT decoding framework, we use the term \emph{latent tokens} to refer to the final-layer hidden state representations produced at designated latent reasoning steps. Concretely, these vectors correspond to the transformer hidden states before the output embedding projection layer is applied, and are used as intermediate continuous representations that condition subsequent generation.

Unlike discrete natural language tokens (e.g., CoT reasoning traces), these latent representations are not explicitly decoded into human-readable text during intermediate steps. Instead, they serve as internal continuous states that guide the model’s subsequent decoding process within the COCONUT framework.

\subsection{Reasoning Paradigms}
We distinguish between two reasoning paradigms commonly studied in recent work:

\textbf{CoT}: $x \rightarrow r \rightarrow y$, where $r$ is an explicit, human-readable reasoning trace. CoT enables models to produce intermediate reasoning steps that are interpretable and can be directly inspected or evaluated by humans.

\textbf{COCONUT}: $x \rightarrow z \rightarrow y$, where $z$ is a sequence of latent tokens that function as placeholders for reasoning. Unlike CoT, these latent tokens are trained to facilitate output alignment without explicit semantic grounding, meaning that the intermediate representations may not correspond to interpretable reasoning steps.

\subsection{Hypotheses}
Based on the above formalization, we formulate two key hypotheses guiding our experimental investigation:

\textbf{H1 (Steering / Controllability)}: If COCONUT latent tokens faithfully encode internal reasoning, then targeted perturbations to these tokens should meaningfully influence the model's final outputs. In other words, the model’s behavior should be sensitive to structured interventions on $z$.

\textbf{H2 (Shortcut / Robustness)}: If COCONUT primarily exploits superficial shortcuts rather than true reasoning, then its predictions are expected to fail under out-of-distribution (OOD) or adversarially designed conditions. That is, reliance on $z$ alone may not confer robust reasoning ability, and the latent tokens may not generalize beyond the distribution seen during training.

\begin{figure*}[h]
    \centering
    \begin{subfigure}{0.32\linewidth}
        \includegraphics[width=\linewidth]{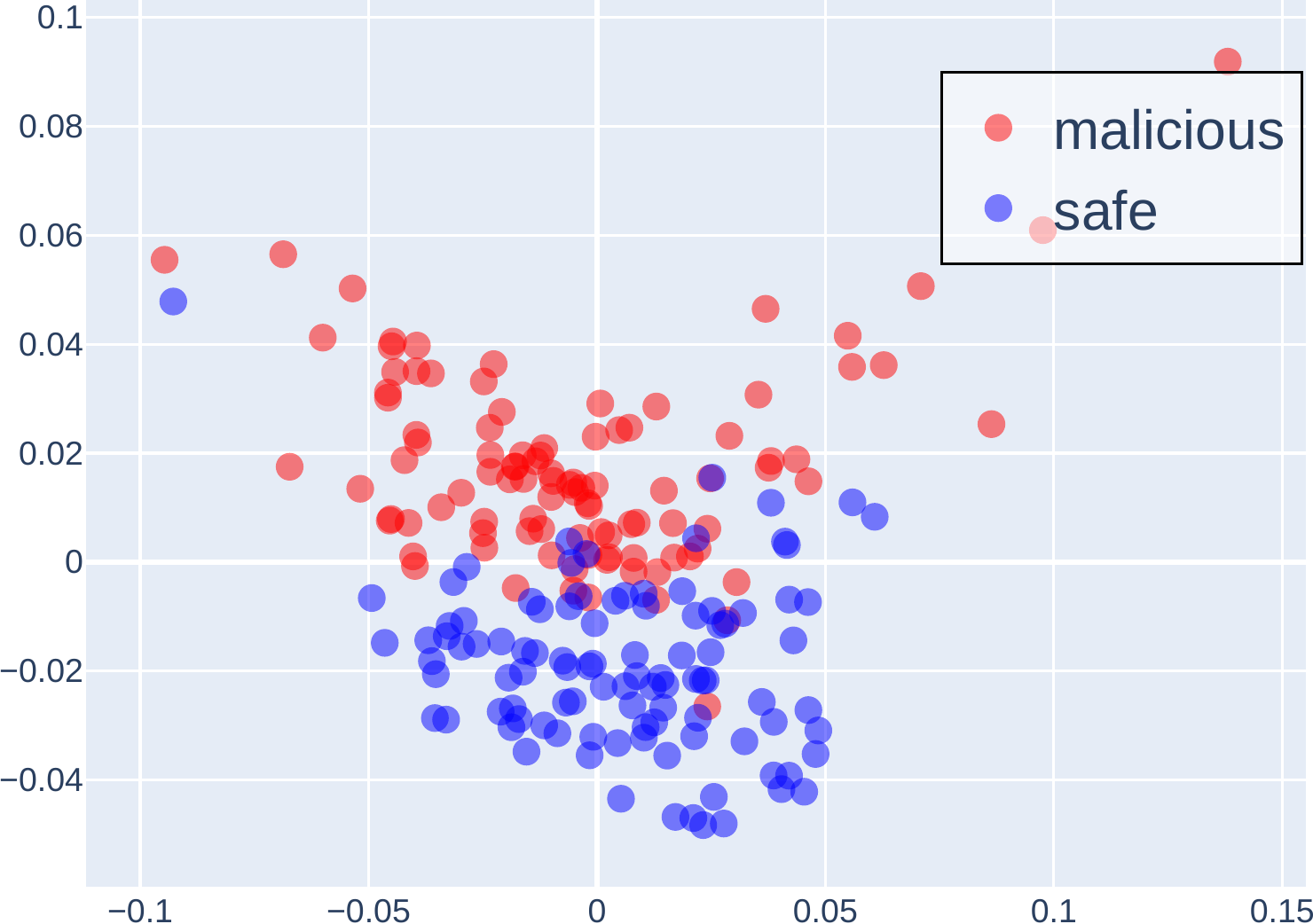}
        \caption{Layer 1}
        \label{fig:layer_1}
    \end{subfigure}
    \hfill
    \begin{subfigure}{0.32\linewidth}
        \includegraphics[width=\linewidth]{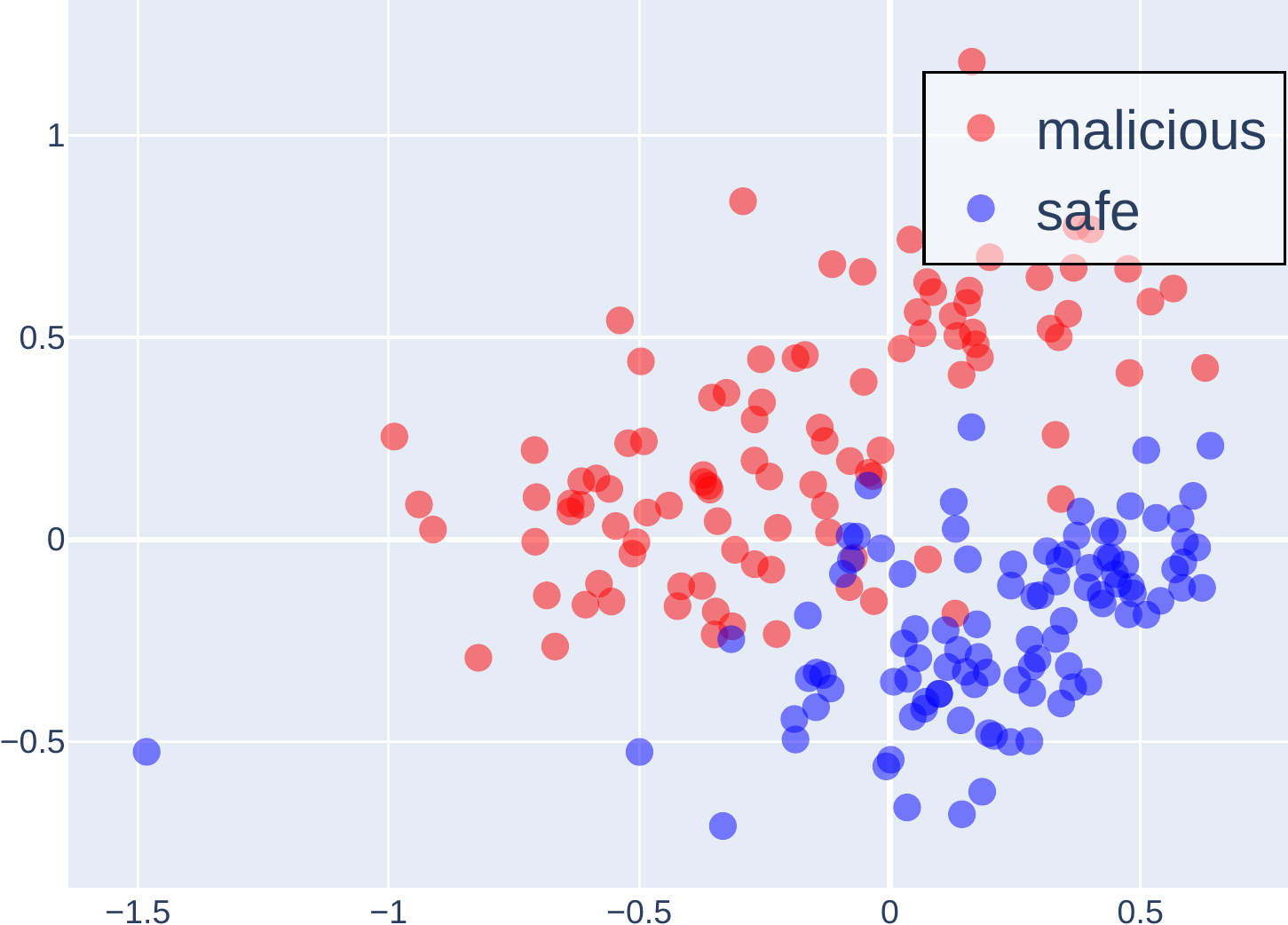}
        \caption{Layer 8}
        \label{fig:layer_8}
    \end{subfigure}
    \hfill
    \begin{subfigure}{0.32\linewidth}
        \includegraphics[width=\linewidth]{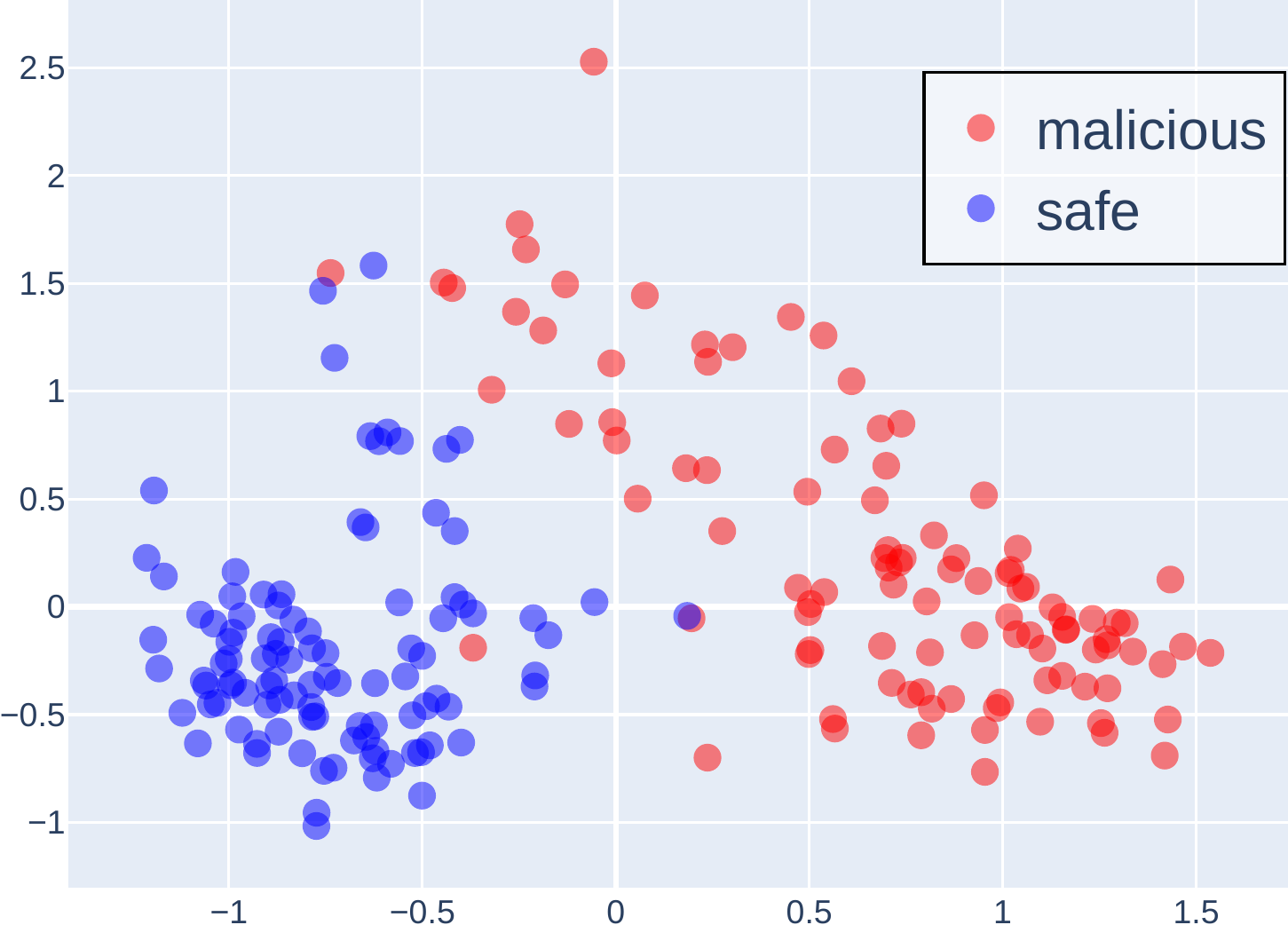}
        \caption{Layer 24}
        \label{fig:layer_24}
    \end{subfigure}
    \caption{PCA Projection of the Last Token Embeddings Across Layers of LLaMA 3 8B Instruct for Malicious and Safe Instructions.}
    \label{fig:reduction_layers}
\end{figure*}

\section{Steering: Method and Experiments}
We first investigate whether COCONUT tokens faithfully represent reasoning by designing steering experiments. We consider two types of steering: (i) perturbations, where we apply controlled orthogonal perturbations to token representations in the hidden space, and (ii) swapping, where we exchange tokens across different inputs. The idea is simple: if these tokens encode meaningful reasoning steps, then steering them in either way should significantly alter model predictions (see Figure~\ref{fig:steering}). 

Importantly, our steering experiments can be viewed as a form of causal intervention, inspired by recent causal interpretability approaches such as activation patching and causal tracing. By directly intervening on the COCONUT tokens while keeping the rest of the model’s computation fixed, we isolate their causal contribution to the final prediction. From this perspective, the extent to which steering alters model outputs provides a direct measure of whether these tokens encode reasoning-relevant information.

\subsection{Method}
Our approach consists of three main components: 
(i) aligning the model’s reasoning behavior via task-specific fine-tuning; 
(ii) preparing latent representations of COCONUT tokens, either by training probes to measure their separability (for perturbation experiments) or by collecting model-generated tokens across the dataset (for swapping experiments); 
and (iii) steering the reasoning process by intervention, where we either apply orthogonal perturbations to the hidden representations, or swap tokens across different samples.

\subsubsection{Probe analysis and token preparation} 
For perturbation experiments, we adopt a linear probing framework to identify whether the model’s latent space encodes separable concept directions. Specifically, for each layer, we train a lightweight linear classifier (probe) on labeled training samples to predict a target concept (e.g., harmful vs. harmless instructions or persona attributes) from intermediate hidden representations. This yields a probe accuracy for each layer, which allows us to identify layers where the concept is well represented and linearly separable.
An example of probing separability in our setting is illustrated in Figure~\ref{fig:reduction_layers}.

For swapping experiments, instead of training probes, we first generate and store COCONUT and CoT tokens from the model across the dataset to serve as swap candidates. 

\subsubsection{Steering via intervention} 
Once probes establish separability (or tokens are collected, for swapping), we steer the reasoning process during generation. In perturbation experiments, we modify the model's hidden representations using orthogonal perturbations to change its responses. This approach is conceptually similar to frameworks such as Safety Concept Activation Vector \citep{xu2024SCAV} and personality-editing approaches \citep{ju2025probing}, but here is used as a causal diagnostic tool to assess whether latent tokens encode reasoning-relevant information.

Specially, given a hidden vector $\mathbf{e} \in \mathbb{R}^d$ at a particular layer, we aim to modify it such that its association with a target concept is reduced, while minimally altering other information encoded in the representation. We first use the corresponding probe to compute the probability $P(\mathbf{e})$ of $\mathbf{e}$ being classified as the target concept. If the probe accuracy at this layer exceeds a threshold $A_0$ (indicating reliable concept separability) and $P(\mathbf{e}) > P_1$, we apply a controlled perturbation to $\mathbf{e}$ to reduce its association with the target concept.

Formally, we seek a perturbation of the form $\tilde{\mathbf{e}} = \mathbf{e} + \epsilon \cdot \mathbf{v}$ such that the modified representation satisfies:
\begin{equation}
P(\tilde{\mathbf{e}}) = P(\mathbf{e} + \epsilon \cdot \mathbf{v}) \leq P_0,
\end{equation}
where $P_0 < P_1$ is a target threshold. Here, $\mathbf{v}$ represents the perturbation direction and is constrained to be a unit vector, while $\epsilon$ controls the magnitude of the perturbation.

To ensure minimal interference with other features in the representation, we solve:
\begin{equation}
\arg\min_{\epsilon, \mathbf{v}} \|\epsilon\|, \text{s.t.} P(\mathbf{e} + \epsilon \cdot \mathbf{v}) \leq P_0, \|\mathbf{v}\|_2 = 1.
\end{equation}

This formulation ensures that the perturbation minimally modifies the hidden representation while achieving the desired change in concept probability. By constraining $\mathbf{v}$ to be a unit vector and applying orthogonal perturbations, we reduce interference with unrelated directions in the latent space, enabling targeted manipulation of specific features while preserving the overall structure of the representation.

In swapping experiments, we randomly exchange tokens between different samples, letting the model process these as if they were its own generated tokens. Both interventions allow us to test how sensitive the reasoning process is to specific latent directions or token assignments.

\subsubsection{Perturbation timing} 
In perturbation experiments, we consider multiple intervention points: (i) Perturbing the embeddings of latent tokens during the COCONUT continuous reasoning process; (ii) Perturbing the embeddings of generated CoT tokens during the explicit CoT reasoning process; (iii) Perturbing the embeddings of all generated tokens. 

\subsection{Experiments}
\label{sec:steering_exp}
\textbf{Datasets.} 
To align reasoning strategies, we first fine-tune the models on the ProntoQA \citep{saparov2022prontoqa} dataset. 
For perturbation experiments, we use two datasets with strong directional tendencies: the AdvBench \citep{zou2023universal} dataset, and the PersonalityEdit \citep{mao2024personalityedit} dataset. For token-swapping experiments, we use the MMLU \citep{hendrycks2021measuring} dataset. 

\textbf{Models.} 
For both perturbation and swap experiments, we conduct studies using four open-source LLMs: LLaMA 3 8B Instruct \citep{llama3modelcard}, LLaMA 2 7B Chat \citep{llama2}, Qwen 2.5 7B Instruct \citep{qwen2.5}, and Falcon 3 7B Instruct \citep{falcon3}, all fine-tuned with full-parameter training. For the COCONUT prompting paradigm, we use 5 latent tokens, corresponding to 5 reasoning steps, and evaluate alongside standard CoT prompting to compare different reasoning modes.


\textbf{Evaluation protocol.} 
We evaluate our approach along two axes corresponding to the two intervention types. 
For perturbation experiments, we measure perturbation effectiveness by perburbation success rate (PSR). It measures the proportion of perturbations that successfully push the model's latent state in the intended target direction. Success is automatically judged by a GPT-4o evaluator, and the prompt used for evaluation is provided in Appendix~\ref{appendix:prompts}.
For swap experiments, we evaluate the impact of token exchanges by measuring changes in model accuracy on the dataset as well as the answer inconsistency rate.

\begin{table}[t]
    \centering
    \scriptsize
    \caption{Perturbation success rates (PSR, \%) on the AdvBench dataset. 
    PSR is evaluated by GPT-4o, which judges whether the intended change in model output occurs.}
    \resizebox{\linewidth}{!}{%
    \begin{tabular}{lcccccc}
    \toprule
    \multirow{2}{*}{Model} & \multicolumn{2}{c}{CoT} & \multicolumn{2}{c}{COCONUT} & \multicolumn{2}{c}{All} \\
    \cmidrule(lr){2-7}
     & Bf & Af & Bf & Af & Bf & Af \\ 
    \midrule
    LLaMA 3 8B   & 0 & 50.00 & 0 &  5.00 & 0 & 100 \\
    LLaMA 2 7B   & 0 & 57.92 & 0 &  0    & 0 & 100 \\
    Qwen 2.5 7B  & 0 & 11.87 & 0 &  9.62 & 0 & 100 \\
    Falcon 3 7B  & 0 & 11.92 & 0 &  0    & 0 & 9.42 \\
    \bottomrule
    \end{tabular}%
    }
    \label{tab:safety}
\end{table}

\begin{table*}[t]
    \centering
    \caption{Perturbation results on the PersonalityEdit dataset. 
Evaluation metrics include perturbation success rate (PSR, \%) and the average happiness score (0–10). 
Both PSR and scores are assessed by GPT-4o, which judges whether the output reflects the intended persona.}
    \resizebox{0.9\linewidth}{!}{%
    \begin{tabular}{cllcccl}
    \toprule
         \multirow{2}{*}{Model}& \multicolumn{2}{c}{CoT}&  \multicolumn{2}{c}{COCONUT}&  \multicolumn{2}{c}{All}\\
         \cmidrule(lr){ 2 - 7 }
 & Before&After& Before& After& Before&After\\\midrule
         LLaMA 3 8B& 26 / 1.81&100 / 9.96&  3 / 0.19&  3.75 / 0.26&  26.25 / 1.87&100 / 10\\
         LLaMA 2 7B& 31.25 / 2.31&46.75 / 4.19&  22 / 1.53&  17.75 / 1.21&  15.75 / 1.11&100 / 10\\
         Qwen 2.5 7B& 8 / 0.55&93.75 / 9.20&  7.5 / 0.50&  9.5 / 0.61&  5.25 / 0.34&100 / 10\\
         Falcon 3 7B& 22 / 1.49&75.25 / 6.69&  7.5 / 0.53&  6.25 / 0.42&  4.25 / 0.27&100 / 10\\ \bottomrule 
    \end{tabular}%
    }
    \label{tab:personality}
\end{table*}

\subsection{Results}
We begin by examining whether latent reasoning tokens in COCONUT can be effectively steered through targeted perturbations.
Table~\ref{tab:safety} reports the perturbation success rates (PSR) on the AdvBench dataset under three perturbation strategies: CoT-only perturbation, COCONUT-only perturbation, and perturbation applied to all tokens. Prior work \citep{xu2024SCAV} has shown that perturbing all tokens can achieve nearly 100\% success rate, which is largely consistent with our findings, except for Falcon 3 7B, where perturbing all tokens yields a PSR of only 9.42\%. This may be due to the stronger safety alignment of Falcon 3 7B, which makes it more resistant to perturbations. Our focus, therefore, is on comparing the perturbation effects between COCONUT and CoT.
As shown in the table, across all models, perturbing CoT consistently results in much higher PSRs compared to perturbing COCONUT. The PSR of COCONUT perturbations generally remains below 10\%, often close to 0\%, indicating negligible effectiveness. In contrast, for LLaMA 3 8B and LLaMA 2 7B, perturbing CoT achieves PSRs of 50\% or higher, suggesting that perturbing CoT can significantly influence the model’s output. Because our perturbations are designed to shift the model’s internal embeddings from unsafe to safe, effectively making it produce valid responses to harmful prompts, it is striking that perturbing CoT succeeds in doing so whereas perturbing COCONUT does not.

To test whether this pattern extends beyond safety steering, we turn to the PersonalityEdit dataset (Table~\ref{tab:personality}), which measures persona-edit success rates and average evaluation scores. Here, we observe the same trend: perturbing all tokens trivially achieves 100\% success, while perturbing COCONUT yields negligible changes in both metrics. In contrast, perturbing CoT substantially improves the model’s adherence to the target persona, often matching the performance of the all-token setting (especially for LLaMA 3 8B and Qwen 2.5 7B).

\begin{table}[t!]
\centering
\caption{Accuracy (\%) and answer inconsistency rate (IR, \%) for the latent token swap experiments on the MMLU dataset. Values are reported as CoT / COCONUT. Here, answer inconsistency rate refers to the percentage of questions whose predicted answers change after token swapping.}
\label{tab:latent_swap}
\resizebox{\columnwidth}{!}{%
\begin{tabular}{lccc}
\toprule
Model & Orig. Acc. & Swapped Acc. & IR \\
\midrule
LLaMA 3 8B  & 62.8 / 60.9 & 43.4 / 61.0 & 52.8 / 17.9 \\
LLaMA 2 7B  & 75.2 / 39.7 & 25.3 / 40.6 & 73.1 / 7.8 \\
Qwen 2.5 7B  & 75.2 / 71.3 & 37.3 / 69.9 & 62.3 / 11.4 \\
Falcon 3 7B  & 70.5 / 74.4 & 23.7 / 71.0 & 74.9 / 23.1 \\
\bottomrule
\end{tabular}
}
\end{table}

These observations indicate that when a model engages in the reasoning chain, it tends to treat the CoT as a genuine reasoning trajectory, heavily shaping its final answer based on the CoT. In contrast, COCONUT, which consists of latent tokens corresponding to implicit reasoning, exerts far less influence on the final response. This suggests that \textbf{models are substantially more likely to regard CoT, rather than COCONUT, as a meaningful component of their reasoning process}.

To further investigate the cause of this insensitivity, we conduct the token-swapping experiment (Table~\ref{tab:latent_swap}). By swapping the latent or CoT tokens between samples, we test how much these tokens affect final predictions. In most cases, before swapping, both COCONUT and CoT achieved accuracies of around 60-70\%. But after swapping, COCONUT’s accuracy remained at a similar level, whereas CoT’s accuracy dropped substantially to around 20-40\%. In terms of inconsistency, COCONUT maintained a low rate of around 10-20\%, while CoT reached 50-70\%, exceeding half of the samples. Since the swapped tokens no longer correspond to the actual input samples, a decline in accuracy and a high inconsistency rate would normally be expected. The fact that COCONUT’s accuracy remains stable, combined with its much lower inconsistency rate, indicates that \textbf{its latent tokens exert very limited influence on the model’s final predictions}.
This suggests that the model does not rely on these latent representations to perform step-by-step reasoning; instead, the reasoning process is likely carried out directly by the base model, with the latent tokens providing at most a weak directional bias.

\section{Shortcut: Method and Experiments}
We next examine whether COCONUT systematically exploits dataset shortcuts. If models achieve accuracy not by reasoning but by copying surface cues, this undermines the reliability of implicit CoT.

\begin{figure*}[t]
    \centering
    \includegraphics[width=\textwidth]{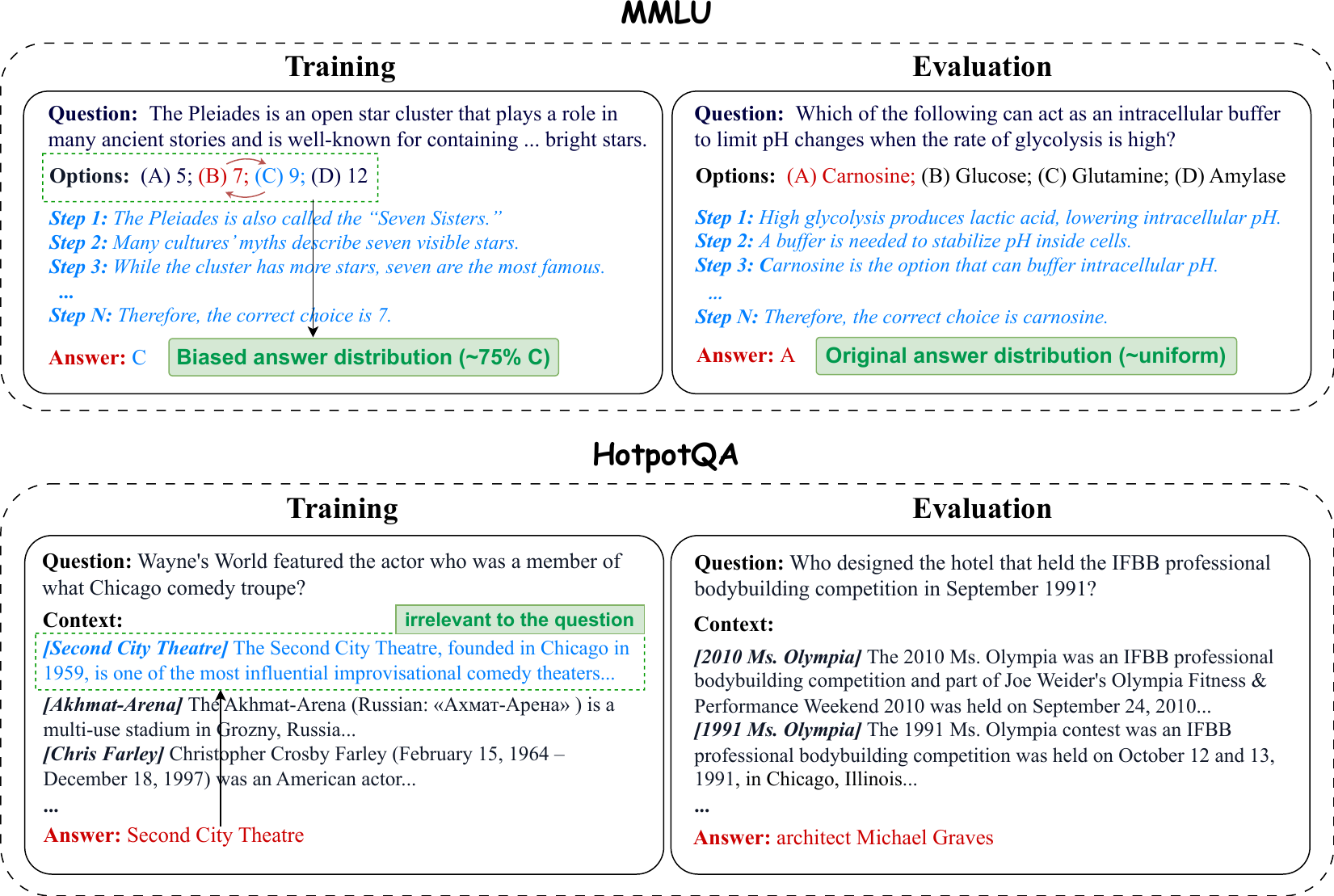}
    \caption{Illustration of the shortcut experiments. Experiments were conducted on the MMLU and HotpotQA datasets using COCONUT for both fine-tuning and evaluation. 
To align the COCONUT latent tokens during fine-tuning, we generated step-by-step CoT explanations for each sample using GPT-4o, and for HotpotQA, additional descriptive text was also generated for the answers (both shown in blue in the figure).}
\end{figure*}

\subsection{Method}
To systematically study shortcut learning in LLMs, we design two types of \textit{shortcut interventions}. 

\textbf{Option manipulation.} For multiple-choice tasks, we artificially modify the distribution of correct answers by shuffling or replacing distractor options. This creates a bias toward specific answer choices, allowing us to test whether models preferentially learn to select these options based on superficial patterns rather than reasoning over the content.

\textbf{Context injection.} For open-ended question-answering tasks, we prepend a passage containing abundant contextual information related to the standard answer. Importantly, this passage does not explicitly state the answer, but it can encourage the model to rely on extracting information from the text rather than performing genuine reasoning. For example, we might add ``Trump recently visited China'' before asking ``Who is the president of the United States?''. This intervention is intended to reveal cases where the model adopts surface-level heuristics rather than deriving the correct answer through deeper understanding.

Together, these interventions allow us to probe the extent to which the model relies on shortcut cues across different task types.

\subsection{Experiments}
\label{sec:shortcut_exp}
\textbf{Datasets and Tasks.} For multiple-choice experiments (\textit{option manipulation}), we use the MMLU \citep{hendrycks2021measuring} dataset. 
For open-ended question-answering (\textit{context injection}), we use the HotpotQA \citep{yang2018hotpotqa} dataset. 

\begin{figure*}[t]  
    \centering
    \begin{subfigure}[b]{0.23\linewidth}
        \includegraphics[width=\linewidth]{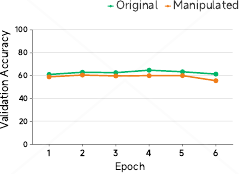}
        \caption{MMLU: validation accuracy}
        \label{fig:shortcut1}
    \end{subfigure}
    \hfill
    \begin{subfigure}[b]{0.23\linewidth}
        \includegraphics[width=\linewidth]{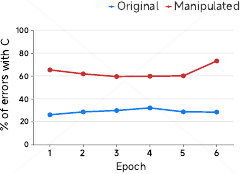}
        \caption{MMLU: fraction of incorrect C choices}
        \label{fig:shortcut2}
    \end{subfigure}
    \hfill
    \begin{subfigure}[b]{0.24\linewidth}
        \includegraphics[width=\linewidth]{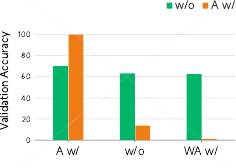}
        \caption{HotpotQA: validation accuracy}
        \label{fig:shortcut3}
    \end{subfigure}
    \hfill
    \begin{subfigure}[b]{0.23\linewidth}
        \includegraphics[width=\linewidth]{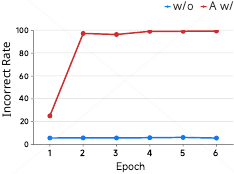}
        \caption{HotpotQA: rate of incorrect shortcut selections}
        \label{fig:shortcut4}
    \end{subfigure}
    \caption{Shortcut experiments on MMLU and HotpotQA.
(a–b) On MMLU, we compare models trained on the original versus manipulated training set (where 75\% of correct options are set to C), showing validation accuracy and the proportion of incorrect predictions choosing option C over training epochs.
(c–d) On HotpotQA, We evaluate models trained with standard answers either with (\textit{A w/}) or without (\textit{w/o}) shortcuts in the training set. Test sets include standard answers with shortcut (\textit{A w/}), without shortcut (\textit{w/o}), and wrong answers with shortcut (\textit{WA w/}). We report validation accuracy (c) and the rate of incorrect predictions selecting the shortcuted incorrect answer (d) over epochs.
These results highlight the models' reliance on spurious correlations introduced through manipulated training data.}
    \label{fig:shortcut_all}
\end{figure*}

\begin{figure*}[t]  
    \centering
    \begin{subfigure}[b]{0.3\linewidth}
        \includegraphics[width=\linewidth]{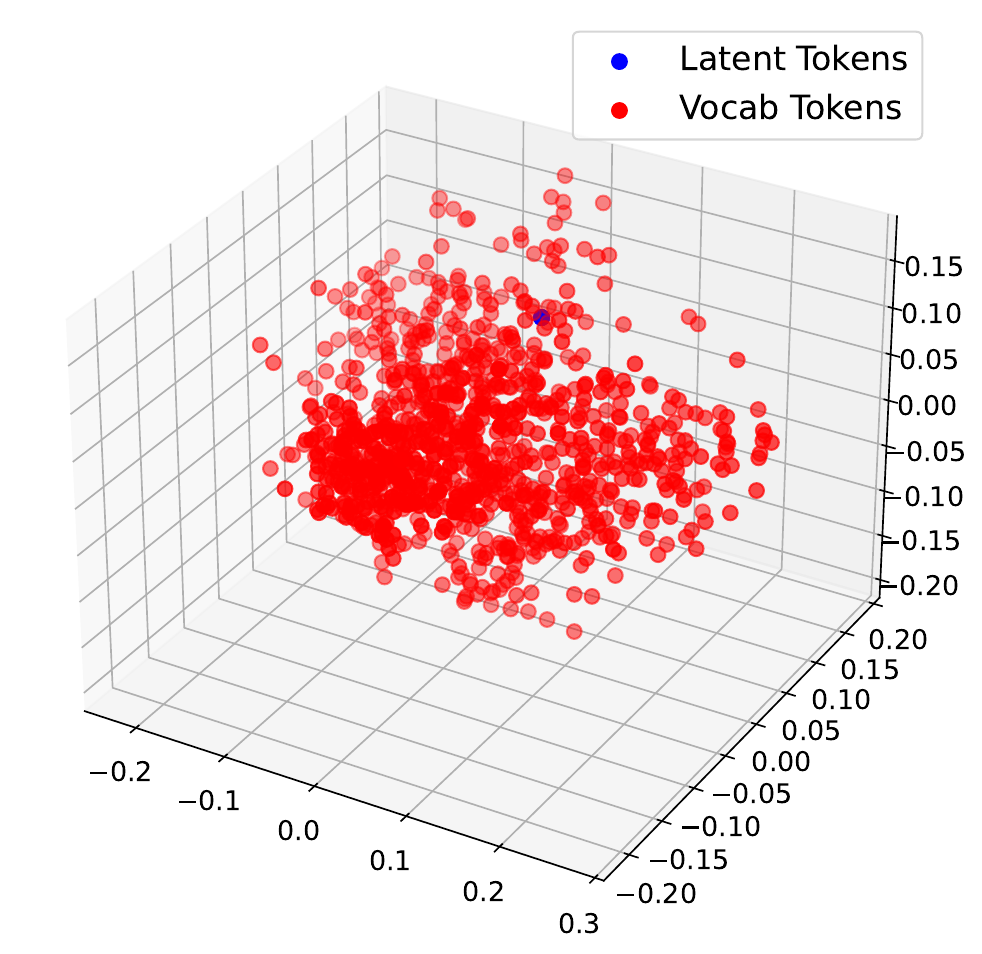}
        \caption{Input embeddings before forward pass}
        \label{fig:pca_input}
    \end{subfigure}
    \hfill
    \begin{subfigure}[b]{0.3\linewidth}
        \includegraphics[width=\linewidth]{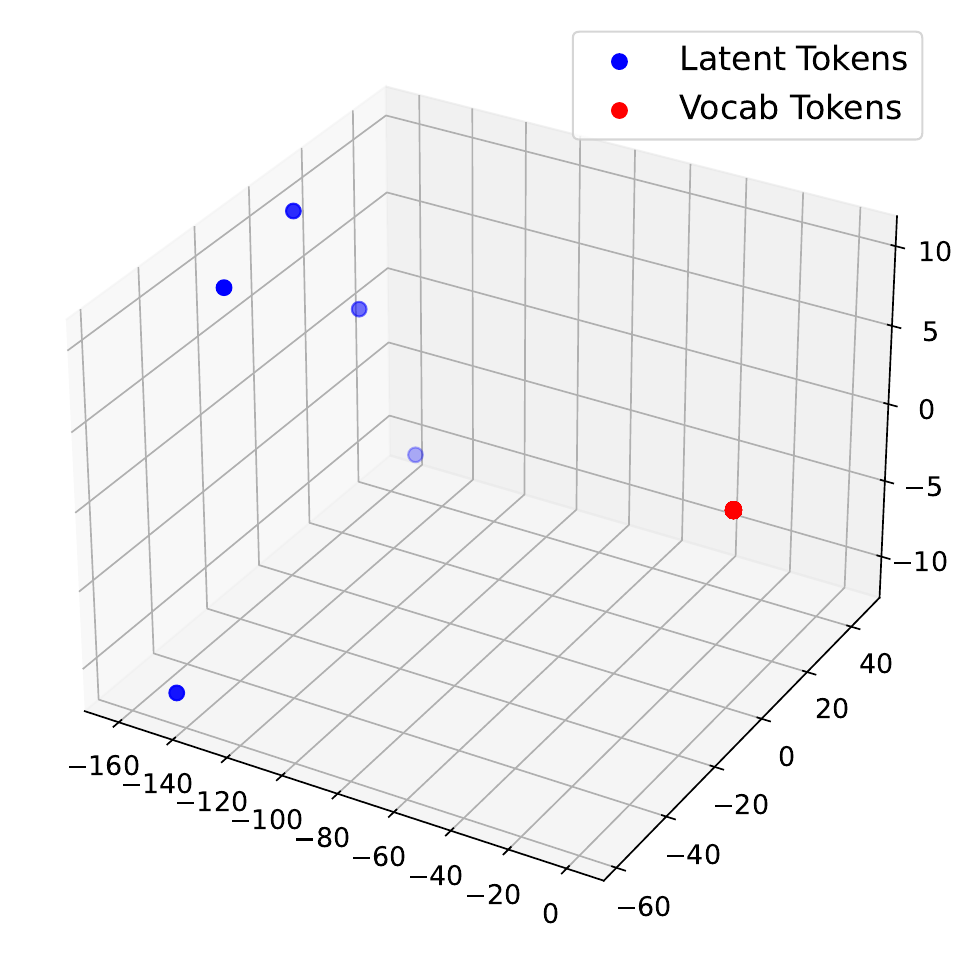}
        \caption{Latent token embeddings after COCONUT reasoning (fine-tuned)}
        \label{fig:pca_ft}
    \end{subfigure}
    \hfill
    \begin{subfigure}[b]{0.3\linewidth}
        \includegraphics[width=\linewidth]{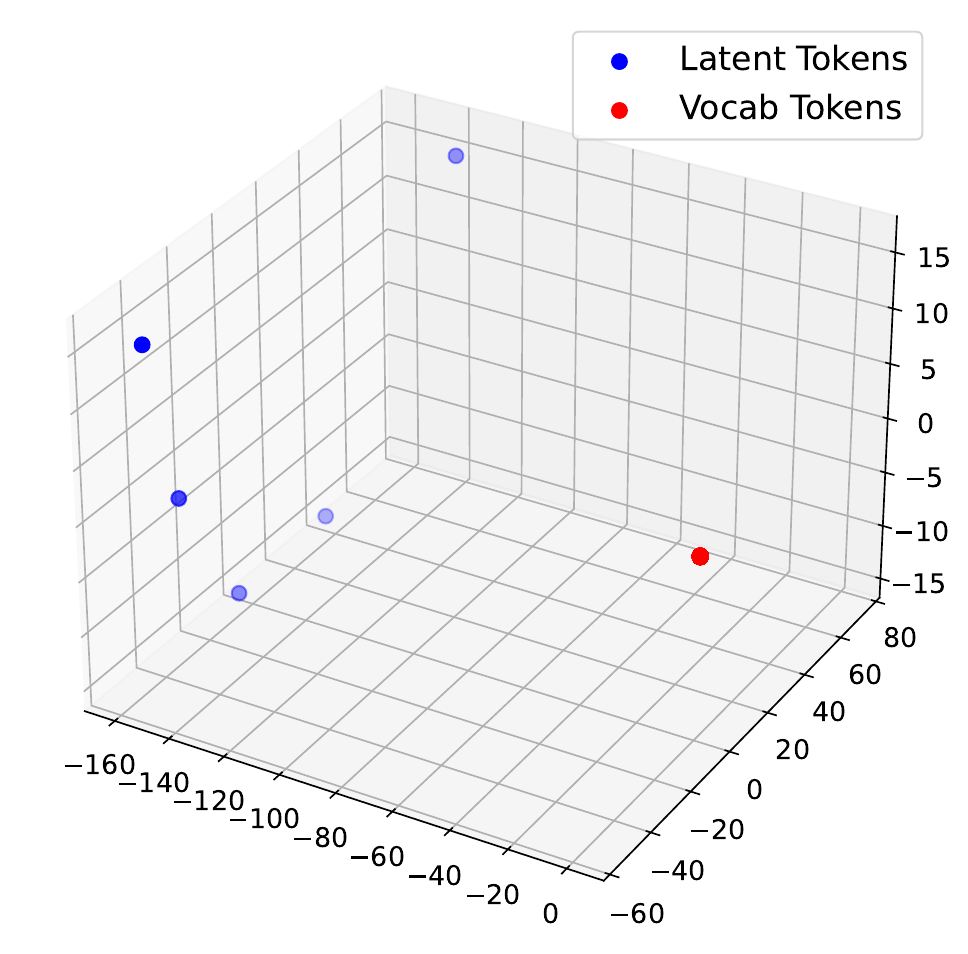}
        \caption{Latent token embeddings after COCONUT reasoning (zero-shot)}
        \label{fig:pca_zero}
    \end{subfigure}
    \caption{3D PCA visualization of latent token embeddings and vocabulary embeddings in LLaMA 3 8B Instruct.}
    \label{fig:pca_all}
\end{figure*}

\textbf{Models and Fine-tuning.} We conduct all experiments with the LLaMA 3 8B Instruct model, chosen for its strong performance on challenging tasks such as MMLU and HotpotQA.
The model is fine-tuned using the COCONUT framework, and we evaluate its performance across training epochs under different experimental conditions.

\textbf{Experimental Design.} For option manipulation, we bias the training set so about 75\% of correct answers are option C, while keeping the test set uniformly distributed. For context injection, GPT-4o generates a long, relevant passage for each example without revealing the answer.
For COCONUT fine-tuning, GPT-4o additionally generates reasoning chains of up to six steps as supervision.

\subsection{Results}
We report the results of the shortcut experiments in Figure~\ref{fig:shortcut_all}. Figure~\ref{fig:shortcut1} and Figure~\ref{fig:shortcut2} present results on the MMLU dataset, examining whether COCONUT amplifies shortcut learning in multiple-choice settings. Figure~\ref{fig:shortcut1} shows that training on a manipulated dataset, where 75\% of correct answers are option C, slightly lowers validation accuracy compared to the balanced dataset. More strikingly, Figure~\ref{fig:shortcut2} shows the rate of incorrect predictions selecting option C rises to about 70\% versus roughly 30\% for the original model, indicating that \textbf{COCONUT fine-tuning induces strong shortcut bias, causing over-reliance on spurious answer patterns rather than genuine task understanding}.

We next move to the open-ended HotpotQA dataset, where shortcuts are injected into the input context instead of answer options (Figures~\ref{fig:shortcut3} and~\ref{fig:shortcut4}).
In Figure~\ref{fig:shortcut3}, we evaluate models trained under two conditions: with shortcuts added to the standard answers and without any shortcuts. Performance is measured on three types of test sets. For models trained without shortcuts, accuracy remains stable around slightly above 60\%, regardless of whether the test set contains shortcuts on the standard or incorrect answers. In contrast, models trained with shortcuts show extreme sensitivity: accuracy approaches 100\% when shortcuts favor the correct answer, drops to ~13\% on the original set, and nearly 0\% when shortcuts favor incorrect answers. This demonstrates a dramatic sensitivity to shortcut manipulation.

To further examine this phenomenon, Figure~\ref{fig:shortcut4} isolates the test condition where shortcuts on incorrect answers. Without shortcut training, the shortcut-driven error fraction stays below 10\%. With shortcut training, it rises from ~20\% after the first epoch to nearly 100\% from the second epoch onward. Since COCONUT gradually introduces latent tokens during training (see Appendix~\ref{appendix:ft}), the first epoch reflects pure CoT reasoning, and subsequent epochs incorporate latent tokens. The sharp increase in shortcut-driven errors after enabling latent tokens suggests that \textbf{even in multi-hop reasoning tasks, COCONUT encourages heavy shortcut reliance rather than genuine reasoning}.

\section{Further Discussion of Latent CoT}
Latent reasoning frameworks like COCONUT are primarily optimized for output alignment, rather than the validity or interpretability of intermediate reasoning steps. Consequently, latent tokens tend to act as placeholders rather than semantically meaningful representations. To further explore this phenomenon, we visualize the latent token embeddings alongside the model's full vocabulary embeddings using 3D PCA (Figure~\ref{fig:pca_all}).

In Figure~\ref{fig:pca_input}, we plot the original input embeddings, including those corresponding to latent tokens, before any forward pass. Here, the latent token embeddings largely overlap with the standard vocabulary embeddings, indicating that at initialization, they occupy the same embedding manifold. In contrast, Figures~\ref{fig:pca_ft} and~\ref{fig:pca_zero} show the embeddings of latent tokens after being processed through the model's COCONUT reasoning steps. Figure~\ref{fig:pca_ft} corresponds to a model fine-tuned on the ProntoQA dataset using the COCONUT paradigm, while Figure~\ref{fig:pca_zero} corresponds to the same reasoning procedure applied without any fine-tuning. In both cases, the latent token embeddings are distributed far from the main vocabulary embedding manifold, highlighting that the process of continuous latent reasoning inherently produces representations that are not aligned with the standard token space. 

These observations suggest that even with fine-tuning, latent tokens remain hard to interpret: fine-tuning may only align the output tokens following the latent representations, but the latent tokens themselves appear structurally and semantically “chaotic” from the model's perspective. Combined with the results of our steering experiments, these findings point to a consistent picture, namely that \textbf{latent tokens primarily serve as placeholders in COCONUT, encoding little directly interpretable information}.

Although COCONUT-style reasoning can sometimes improve task performance, our previous experiments indicate these gains may stem from exploiting \textit{shortcuts} rather than genuine reasoning. We hypothesize that during COCONUT-style fine-tuning, the model is optimized via gradient descent to minimize training loss as quickly as possible. This optimization process naturally encourages reliance on easy-to-learn shortcut features—spurious yet highly predictive signals that improve performance on the training data without reflecting true reasoning ability. In this setting, latent tokens are more likely to capture and amplify such shortcut signals, rather than encoding a causal, step-by-step reasoning process. This perspective explains why perturbing latent tokens has little effect on downstream reasoning behavior: the model does not depend on these representations to perform reasoning, but instead relies on its base computation. Consistent with this view, our shortcut experiments further show that latent tokens tend to capture superficial correlations rather than structured reasoning patterns. Future work could formalize this hypothesis using techniques such as gradient attribution or information bottlenecks to better characterize the information content of latent tokens.

\section{Conclusion}
In this work, we present the first systematic evaluation of the faithfulness of latent reasoning in LLMs. Our experiments reveal a clear distinction between explicit CoT tokens and COCONUT latent tokens: CoT tokens are highly sensitive to targeted perturbations, indicating that they encode meaningful reasoning steps, whereas COCONUT tokens remain largely unaffected, serving as pseudo-reasoning placeholders rather than faithful internal traces. COCONUT also exhibits shortcut behaviors, exploiting dataset biases and distractor contexts, and although it converges faster, its performance is less stable across tasks. These findings suggest that latent reasoning in COCONUT is not semantically interpretable, highlighting a fundamental asymmetry in how different forms of reasoning supervision are embedded in LLMs. Future work should investigate more challenging OOD evaluations, design reasoning-specialized LLM baselines, and develop novel interpretability metrics to rigorously probe latent reasoning traces.

\section*{Limitations}
Our work has several limitations.
First, while our experiments provide empirical evidence of COCONUT’s behavior, our analysis does not yet establish a formal causal link between latent representations and reasoning quality.
Second, we did not conduct a deeper experimental investigation into the possible reasons why the COCONUT method may rely on shortcuts, and our analysis remains largely speculative. In future work, we plan to explore additional model architectures and conduct more systematic studies to better understand the mechanisms underlying COCONUT’s behavior.

\section*{Ethical Statement}
Our study conducts experiments on LLMs using publicly available datasets, including ProntoQA \citep{saparov2022prontoqa}, MMLU \citep{hendrycks2021measuring}, AdvBench \citep{zou2023universal}, PersonalityEdit \citep{mao2024personalityedit}, and HotpotQA \citep{yang2018hotpotqa}. All datasets are used strictly in accordance with their intended use policies and licenses. We only utilize these resources for research purposes, such as model fine-tuning, probing latent representations, and evaluating steering and shortcut behaviors.

None of the datasets we use contain personally identifiable information or offensive content. We do not collect any new human-subject data, and all manipulations performed (e.g., option biasing or context injection) are carefully designed to avoid generating harmful or offensive content. Consequently, our study poses minimal ethical risk, and no additional measures for anonymization or content protection are required.

Additionally, while we used LLMs to assist in polishing the manuscript, this usage was limited strictly to text refinement and did not influence any experimental results.

\section*{Acknowledgement}
This work was supported by the Joint Funds of the National Natural Science Foundation of China (Grant No. U21B2020) and the National Natural Science Foundation of China (Grant No. 62506230).

\bibliography{anthology, custom}

\clearpage
\appendix

\section{Fine-tuning with COCONUT}
\label{appendix:ft}
All fine-tuning performed on COCONUT in our experiments follows the stepwise procedure proposed in the original COCONUT paper. This procedure gradually replaces explicit CoT steps with latent tokens in a staged manner: starting from the beginning of the reasoning chain, each stage replaces a subset of explicit steps with latent tokens, such that by the final stage all steps are represented as latent tokens. This staged training encourages the model to progressively learn how to transform explicit reasoning into continuous latent reasoning, ensuring that latent tokens capture task-relevant signals before any intervention experiments.

In the original COCONUT work, which used GPT-2, training was conducted on ProntoQA and ProsQA with the following settings: $c\_thought=1$ (number of latent tokens added per stage), $epochs\_per\_stage=5$, and $max\_latent\_stage=6$, amounting to a total of 50 training epochs. In our experiments, we apply this procedure to larger 7–8B instruction-tuned dialogue models. Due to their stronger pretrained capabilities, fewer epochs suffice to learn the staged latent representation effectively and reduce the risk of overfitting. Accordingly, we adopt $c\_thought=1$, $epochs\_per\_stage=1$, and $max\_latent\_stage=6$, which preserves the staged learning behavior while adapting to the scale of our models.

\section{Reproduction of COCONUT Baseline Results}

To validate that the reported performance of COCONUT is not driven by implementation discrepancies, we reproduce key experiments from the original paper under the reported settings.

Specifically, we re-run the original COCONUT experiments using GPT-2-based architectures, following the datasets, evaluation metrics, and training pipeline described in the original work. We adhere to the original implementation as closely as possible, without modifying hyperparameters unless explicitly specified.

The reproduced results are shown in Table~\ref{tab:repro}. Overall, they are consistent with the trends reported in the original paper, indicating that the performance patterns of COCONUT are stable under the original experimental setup.

\begin{table}[h]
\centering
\resizebox{\columnwidth}{!}{
\begin{tabular}{lccc}
\hline
Method & GSM8K & ProntoQA & ProsQA \\
\hline
CoT (Original) & 42.9 & 98.8 & 77.5 \\
CoT (Reproduced) & 42.0 & 100.0 & 77.7 \\
COCONUT (Original) & 34.1 & 99.8 & 97.0 \\
COCONUT (Reproduced) & 34.2 & 99.9 & 96.3 \\
\hline
\end{tabular}
}
\caption{Reproduction of COCONUT baseline results under original settings.}
\label{tab:repro}
\end{table}

\section{Training Setups}
All fine-tuning experiments are performed using a batch size of 128, a learning rate of $1\times10^{-5}$, weight decay of 0.01, and the AdamW optimizer. Training is conducted with bfloat16 precision. 

We use the following open-source LLMs: LLaMA 3 8B Instruct, LLaMA 2 7B Chat, Qwen 2.5 7B Instruct, and Falcon 7B Instruct. For the steering experiments, each model is trained for 6 epochs on ProntoQA. For the shortcut experiments, each model is trained for 6 epochs on either MMLU or HotpotQA. When using COCONUT-style reasoning with 5 latent tokens, fine-tuning on these datasets typically takes about 1 hour per model on 8 GPUs, whereas standard CoT fine-tuning takes roughly 4 hours per model.

\subsection{Implementation Details}
We rely on the Hugging Face Transformers library \citep{wolf-etal-2020-transformers} for model loading, tokenization, and training routines. All models are loaded from their corresponding Hugging Face checkpoints, and we use the default tokenizer configurations unless otherwise specified. For evaluation, we compute standard metrics using scikit-learn. No additional preprocessing packages (e.g., NLTK, SpaCy) were required beyond standard tokenization.

\section{Dataset Details}
\subsection{Datasets for Steering Experiments}
\label{app:steering_dataset}
We provide additional details about the datasets used in Section~\ref{sec:steering_exp}.

\paragraph{AdvBench.} 
The AdvBench dataset contains 520 samples. 
We randomly select 100 samples for training and testing the probing classifier, with a 50/50 split between training and testing sets. 
Within each split, the number of \textit{malicious} and \textit{safe} samples is balanced. 
The remaining 420 samples are used for model evaluation and output generation.

\paragraph{PersonalityEdit.} 
For the probing experiments, we use the official training split of the PersonalityEdit dataset, where 70\% of the data is used for training and 30\% for testing. 
Both splits are balanced between the two personality polarities. 
For model output evaluation, we use the \texttt{dev} and \texttt{test} splits combined, again maintaining equal proportions of the two polarities.
Since the dataset mainly consists of questions asking for the model’s opinions on various topics, we introduce polarity by modifying the prompt—for example, by appending the instruction \textit{``Please answer with a very happy and cheerful tone''} to construct the ``happy'' and ``neutral'' variants.

\paragraph{MMLU.} 
For token-swapping experiments, we use 1,000 randomly sampled examples from the \texttt{test} split of the MMLU dataset. 
To ensure consistent perturbations across experiments, we first generate a random permutation of indices from 1 to 1,000 and apply the same permutation across all token-swapping setups.

\subsection{Datasets for Shortcut Experiments}
\label{app:shortcut_dataset}
This section provides additional details about the datasets used in Section~\ref{sec:shortcut_exp}.

\paragraph{MMLU.} 
For multiple-choice experiments (\textit{option manipulation}), we use the full \texttt{all} split of the MMLU dataset. 
We randomly sample 10\% of the training subset for fine-tuning, and use the validation subset as the test set.

\paragraph{HotpotQA.} 
For open-ended question answering (\textit{context injection}), we randomly sample 10\% of the HotpotQA training data for fine-tuning, and select 3{,}000 examples from the validation split for evaluation.

\section{Prompts}
\label{appendix:prompts}
We used different prompt templates depending on the experiment type:

\subsection{Perturbation Experiments}
For perturbation experiments, prompts were designed to elicit either explicit CoT reasoning steps or continuous COCONUT latent tokens, consistent with the fine-tuning setup. This ensures that perturbations can be meaningfully evaluated. 

Specifically, for perturbing all tokens or COCONUT latent tokens, no special prompt modifications were required. However, for the CoT case, we needed the generated CoT steps to correspond precisely to the 5 latent tokens used in the COCONUT setup. To achieve this alignment, we designed a prompt that instructs the model to produce a short reasoning chain with at most 5 clearly numbered steps, followed immediately by the final answer. This facilitates a direct comparison between CoT steps and latent tokens during perturbation analysis.

\begin{promptbox}

First, generate a short reasoning chain-of-thought (at most 5 steps).

Number each step explicitly as ``1.'', ``2.'', ``3.'', etc.

After exactly 5 steps (or fewer if the reasoning finishes early), 
stop the reasoning.

Then, immediately continue with the final answer, starting with ``\#''.

\end{promptbox}

\subsection{Swap Experiments} 
For swap experiments, prompts were designed primarily to standardize the output format, ensuring consistent generation across MMLU samples and facilitating accurate measurement of model accuracy after token exchanges. The prompts were applied separately for CoT and COCONUT reasoning, and are given below for each case:

\begin{promptbox}

You are a knowledgeable assistant.

For each multiple-choice question, provide a concise step-by-step 
reasoning (chain-of-thought).

Number each step starting from 1, using the format ``1.'', ``2.'', etc.

Use at most 5 steps.

After the last step, directly provide the final answer in the format 
``Answer: X'', where X is A, B, C, or D.

Keep each step brief and focused.

\end{promptbox}

\begin{promptbox}

You are a knowledgeable expert.

Please answer the following multiple-choice question correctly.

Do not output reasoning or explanation.

Only respond in the format: ``Answer: X'' where X is one of 
A, B, C, or D.

\end{promptbox}

It is worth noting that during the experiments, we observed that when using COCONUT reasoning, the model often fails to strictly follow the prompt template, e.g., the expected format "Answer: X". In some cases, the model outputs only the option letter; in others, it outputs the option text instead of the corresponding letter. To standardize the outputs, we employed GPT-4o to extract the intended option from the raw COCONUT outputs using the following prompt:

\begin{promptbox}

You are given a multiple-choice question with four options 
(A, B, C, D), and a raw model output that may be noisy or 
unstructured.

Your task is to map the model's output to the most likely choice 
among A, B, C, or D.

\textbf{Instructions:}

1. Read the question and the four answer choices carefully.

2. Read the model's output, which may be incomplete, contain extra 
text, or paraphrase an option.

3. Decide which option (A/B/C/D) the model most likely intended.

4. If the model's output cannot be clearly mapped to any choice, 
output ``0''.

5. Output ONLY one of: A, B, C, D, or 0. Do not output explanations.
\end{promptbox}

\subsection{Shortcut Experiments} 
In this set of experiments, we fine-tuned the model with the COCONUT method on both the MMLU and HotpotQA datasets. Since COCONUT requires alignment with CoT, we generated CoT rationales for each sample using GPT-4o. 
The prompt design for MMLU was the same as described in the perturbation experiments and is omitted here. 
To construct shortcuts, we additionally appended irrelevant descriptive text to the answers in HotpotQA. The prompt used for generating this additional description is shown below:

\begin{promptbox}

You are given a pair of data:
\begin{itemize}
    \setlength{\itemsep}{1pt}
    \setlength{\parskip}{0pt}
    \setlength{\parsep}{0pt}
    \item A hidden question.
    \item Its answer (a noun).
\end{itemize}

Your task is to generate a descriptive passage of no fewer than 
400 words, focusing on the given answer (the noun) as the subject 
of description.

\medskip
\textbf{Requirements:}
\begin{enumerate}
    \setlength{\itemsep}{2pt}
    \setlength{\parskip}{0pt}
    \setlength{\parsep}{0pt}

    \item The passage must be relevant to the answer (the noun) and 
    explore it in depth. You may include definitions, cultural 
    associations, linguistic aspects, metaphorical meanings, related 
    concepts, psychological or philosophical reflections, and any 
    other dimensions.

    \item Do \textbf{not} mention, describe, or imply any knowledge 
    that would directly reveal or be connected to the given question. 
    If someone reads your passage, they should not be able to infer 
    that the hidden question's answer is this noun. In other words, 
    the text must describe the answer in depth without exposing its 
    role as the solution to the hidden question.

    \item The passage should be coherent, detailed, and contain at 
    least 400 words.
\end{enumerate}

\end{promptbox}

\section{Additional Analysis on Text Fluency}
To further examine the impact of COCONUT reasoning on text generation quality, we compute the perplexity of model outputs from the experiments described in Section~\ref{sec:steering_exp}. Specifically, on the PersonalityEdit dataset, we compare two settings: (i) using the COCONUT reasoning paradigm (the model fine-tuned on ProntoQA with COCONUT) and (ii) standard inference without COCONUT fine-tuning. As shown in Table~\ref{tab:ppl}, COCONUT reasoning yields substantially higher perplexity, indicating that it can degrade the fluency or coherence of generated text. Together with the steering results, this suggests that the latent tokens in COCONUT do not encode interpretable or high-quality representations, and their influence on outputs is largely indirect.

\begin{table}[t]
\centering
\caption{Average perplexity (PPL) with and without COCONUT reasoning.}
\label{tab:ppl}
\resizebox{\columnwidth}{!}{%
\begin{tabular}{lcccc}
\toprule
 & LLaMA 3 8B & LLaMA 2 7B & Qwen 2.5 7B & Falcon 3 7B \\
\midrule
COCONUT & \textbf{238.2307} & 9.7098 & \textbf{25.4974} & \textbf{57.1146} \\
Vanilla & 11.1525 & \textbf{10.0651} & 14.2427 & 16.5557 \\
\bottomrule
\end{tabular}%
}
\end{table}


\end{document}